\pdfoutput = 1

\documentclass[11pt,a4paper]{article}
\usepackage[hyperref]{acl2018}
\usepackage{times}
\usepackage{latexsym}
\usepackage{amsfonts}
\usepackage{amsmath}
\usepackage{url}
\usepackage{bbm}

\usepackage{xcolor}
\usepackage{array}
\usepackage{tabulary}
\usepackage{graphicx}
\usepackage{algorithm,algorithmicx,algpseudocode}
\usepackage{hyperref,etoolbox}
\usepackage{mathtools, nccmath}
\usepackage{xspace}
\usepackage{amsfonts}
\usepackage{amssymb}
\usepackage{latexsym}

\makeatletter
\patchcmd{\ALG@step}{\addtocounter{ALG@line}{1}}{\refstepcounter{ALG@line}}{}{}
\newcommand{\ALG@lineautorefname}{Line}
\makeatother

\makeatletter
\def\BState{\State\hskip-\ALG@thistlm}
\makeatother

\newenvironment{myquote}[1]%
  {\list{}{\leftmargin=#1\rightmargin=#1}\item[]}%
  {\endlist}

\aclfinalcopy 
\def\aclpaperid{308} 

\newcommand{\expect}[1]{\ensuremath{\underset{#1}{\mathbb{E}}\xspace}}

\newcommand\samethanks[1][\value{footnote}]{\footnotemark[#1]}

\newcommand{\focus}[1]{\underline{#1}}

\title{No Metrics Are Perfect: \\ Adversarial Reward Learning for Visual Storytelling}

\author{Xin Wang\thanks{\hspace{0.2cm} Equal contribution}~,~ Wenhu Chen\samethanks ~,~ Yuan-Fang Wang~,~ William Yang Wang \\
  University of California, Santa Barbara \\
  {\tt \{xwang,wenhuchen,yfwang,william\}@cs.ucsb.edu}}

\date{}

\begin{document}
\maketitle
\begin{abstract}
Though impressive results have been achieved in visual captioning, the task of generating abstract stories from photo streams is still a little-tapped problem. Different from captions, stories have more expressive language styles and contain many imaginary concepts that do not appear in the images. Thus it poses challenges to behavioral cloning algorithms. Furthermore, due to the limitations of automatic metrics on evaluating story quality, reinforcement learning methods with hand-crafted rewards also face difficulties in gaining an overall performance boost. Therefore, we propose an Adversarial REward Learning (AREL) framework to learn an implicit reward function from human demonstrations, and then optimize policy search with the learned reward function. Though automatic evaluation indicates slight performance boost over state-of-the-art (SOTA) methods in cloning expert behaviors, human evaluation shows that our approach achieves significant improvement in generating more human-like stories than SOTA systems.\footnote{Code is released at \small\url{https://github.com/littlekobe/AREL}}
\end{abstract}

\section{Introduction}

\begin{figure}[!t]
\begin{center}
\includegraphics[width=0.5\textwidth]{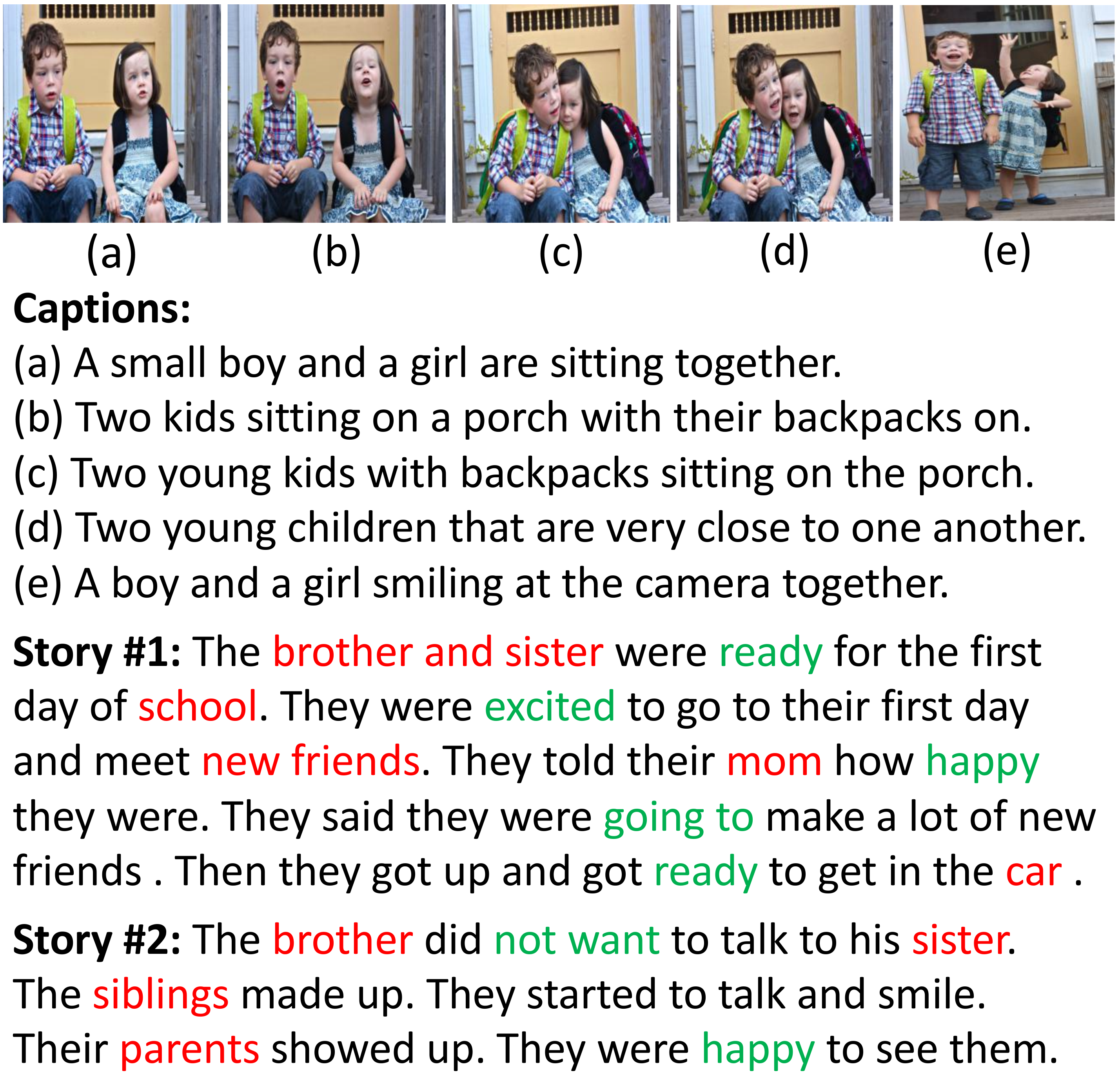}  
\end{center}
   \caption{An example of visual storytelling and visual captioning. Both captions and stories are shown here: each image is captioned with one sentence, and we also demonstrate two diversified stories that match the same image sequence.}
\label{fig:intro}
\end{figure}

Recently, increasing attention has been focused on visual captioning~\cite{chen2015microsoft,chen2016bootstrap,Xu:CVPR16,wang2018watch}, which aims at describing the content of an image or a video. 
Though it has achieved impressive results, its capability of performing human-like understanding is still restrictive. 
To further investigate machine's capabilities in understanding more complicated visual scenarios and composing more structured expressions, visual storytelling~\cite{huang2016visual} has been proposed.
Visual captioning is aimed at depicting the concrete content of the images, and its expression style is rather simple. In contrast, visual storytelling goes one step further: it summarizes the idea of a photo stream and tells a story about it. 
\autoref{fig:intro} shows an example of visual captioning and visual storytelling. We have observed that stories contain rich \textbf{emotions} (\textit{excited, happy, not want}) and \textbf{imagination} (\textit{siblings, parents, school, car}). It, therefore, requires the capability to associate with concepts that do not explicitly appear in the images. Moreover, stories are more \textbf{subjective}, so there barely exists standard templates for storytelling. As shown in \autoref{fig:intro}, the same photo stream can be paired with diverse stories, different from each other. This heavily increases the evaluation difficulty.  

So far, prior work for visual storytelling~\cite{huang2016visual,yu-bansal-berg:2017:EMNLP2017} is mainly inspired by the success of visual captioning. 
Nevertheless, because these methods are trained by maximizing the likelihood of the observed data pairs, they are restricted to generate simple and plain description with limited expressive patterns. In order to cope with the challenges and produce more human-like descriptions, ~\citet{rennie2016self} have proposed a reinforcement learning framework. However, in the scenario of visual storytelling, the common reinforced captioning methods are facing great challenges since the hand-crafted rewards based on string matches are either too biased or too sparse to drive the policy search. For instance, we used the METEOR~\cite{banerjee2005meteor} score as the reward to reinforce our policy and found that though the METEOR score is significantly improved, the other scores are severely harmed. Here we showcase an adversarial example with an average METEOR score as high as 40.2:
\begin{myquote}{0.2in}
\textit{We had a great time to have a lot of the. They were to be a of the. They were to be in the. The and it were to be the. The, and it were to be the}. 
\end{myquote}
Apparently, the machine is gaming the metrics. Conversely, when using some other metrics (\textit{e.g.} BLEU, CIDEr) to evaluate the stories, we observe an opposite behavior: many relevant and coherent stories are receiving a very low score (nearly zero). 

In order to resolve the strong bias brought by the hand-coded evaluation metrics in RL training and produce more human-like stories, we propose an Adversarial REward Learning (AREL) framework for visual storytelling. We draw our inspiration from recent progress in inverse reinforcement learning~\cite{ho2016generative,finn2016connection,fu2017learning} and propose the AREL algorithm to learn a more intelligent reward function. Specifically, we first incorporate a Boltzmann distribution to associate reward learning with distribution approximation, then design the adversarial process with two models -- a \textbf{policy model} and a \textbf{reward model}. The policy model performs the primitive actions and produces the story sequence, while the reward model is responsible for learning the implicit reward function from human demonstrations. The learned reward function would be employed to optimize the policy in return. 

For evaluation, we conduct both automatic metrics and human evaluation but observe a poor correlation between them. Particularly, our method gains slight performance boost over the baseline systems on automatic metrics; human evaluation, however, indicates significant performance boost. Thus we further discuss the limitations of the metrics and validate the superiority of our AREL method in performing more intelligent understanding of the visual scenes and generating more human-like stories.

Our main contributions are four-fold:
\begin{itemize}
\item We propose an adversarial reward learning framework and apply it to boost visual story generation. 
\item We evaluate our approach on the Visual Storytelling (VIST) dataset and achieve the state-of-the-art results on automatic metrics.
\item We empirically demonstrate that automatic metrics are not perfect for either training or evaluation.
\item We design and perform a comprehensive human evaluation via Amazon Mechanical Turk, which demonstrates the superiority of the generated stories of our method on relevance, expressiveness, and concreteness.
\end{itemize}

\section{Related Work}
\paragraph{Visual Storytelling}
Visual storytelling is the task of generating a narrative story from a photo stream, which requires a deeper understanding of the event flow in the stream. \citet{park2015expressing} has done some pioneering research on storytelling. \citet{ijcai2017-chen-multimodal} proposed a multimodal approach for storyline generation to produce a stream of entities instead of human-like descriptions. Recently, a more sophisticated dataset for visual storytelling (VIST) has been released to explore a more human-like understanding of grounded stories~\cite{huang2016visual}. \citet{yu-bansal-berg:2017:EMNLP2017} proposes a multi-task learning algorithm for both album summarization and paragraph generation, achieving the best results on the VIST dataset. But these methods are still based on behavioral cloning and lack the ability to generate more structured stories.

\paragraph{Reinforcement Learning in Sequence Generation}
Recently, reinforcement learning (RL) has gained its popularity in many sequence generation tasks such as machine translation~\cite{bahdanau2016actor}, visual captioning~\cite{Ren_etal_CVPR_17,wang2018video}, summarization~\cite{paulus2017deep,chen2018generative}, etc. 
The common wisdom of using RL is to view generating a word as an action and aim at maximizing the expected return by optimizing its policy. As pointed in~\cite{ranzato2015sequence}, traditional maximum likelihood algorithm is prone to exposure bias and label bias, while the RL agent exposes the generative model to its own distribution and thus can perform better. But these works usually utilize hand-crafted metric scores as the reward to optimize the model, which fails to learn more implicit semantics due to the limitations of automatic metrics.

\paragraph{Rethinking Automatic Metrics}
Automatic metrics, including BLEU~\cite{papineni2002bleu}, CIDEr~\cite{vedantam2015cider}, METEOR~\cite{banerjee2005meteor}, and ROUGE~\cite{lin2004rouge}, have been widely applied to the sequence generation tasks. Using automatic metrics can ensure rapid prototyping and testing new models with fewer expensive human evaluation. However, they have been criticized to be biased and correlate poorly with human judgments, 
especially in many generative tasks like response generation~\cite{lowe2017towards,liu2016not}, dialogue system~\cite{bruni2017adversarial} and machine translation~\cite{callison2006re}. 
The naive overlap-counting methods are not able to reflect many semantic properties in natural language, such as coherence, expressiveness, etc. 

\paragraph{Generative Adversarial Network}
Generative adversarial network (GAN)~\cite{goodfellow2014generative} is a very popular approach for estimating intractable probabilities, which sidestep the difficulty by alternately training two models to play a min-max two-player game:
\begin{align*}
\min_{D}\max_{G} \expect{x \sim p_{data}}[\log D(x)] + \expect{z \sim p_{z}}[\log D(G(z))] ~,
\end{align*}
where $G$ is the generator and $D$ is the discriminator, and $z$ is the latent variable. Recently, GAN has quickly been adopted to tackle discrete problems~\cite{yu2017seqgan,Dai_2017_ICCV,wang2018show}. The basic idea is to use Monte Carlo policy gradient estimation~\cite{williams1992simple} to update the parameters of the generator.     

\paragraph{Inverse Reinforcement Learning}
Reinforcement learning is known to be hindered by the need for an extensive feature and reward engineering, especially under the unknown dynamics. Therefore, inverse reinforcement learning (IRL) has been proposed to infer expert's reward function. Previous IRL approaches include maximum margin approaches~\cite{abbeel2004apprenticeship,ratliff2006maximum} and probabilistic approaches~\cite{ziebart2010modeling,ziebart2008maximum}. Recently, adversarial inverse reinforcement learning methods provide an efficient and scalable promise for automatic reward acquisition~\cite{ho2016generative,finn2016connection,fu2017learning,henderson2017optiongan}. These approaches utilize the connection between IRL and energy-based model and associate every data with a scalar energy value by using Boltzmann distribution $p_{\theta}(x) \propto \exp (-E_{\theta}(x))$. Inspired by these methods, we propose a practical AREL approach for visual storytelling to uncover a robust reward function from human demonstrations and thus help produce human-like stories.

\begin{figure}
\begin{center}
\includegraphics[width=1\linewidth]{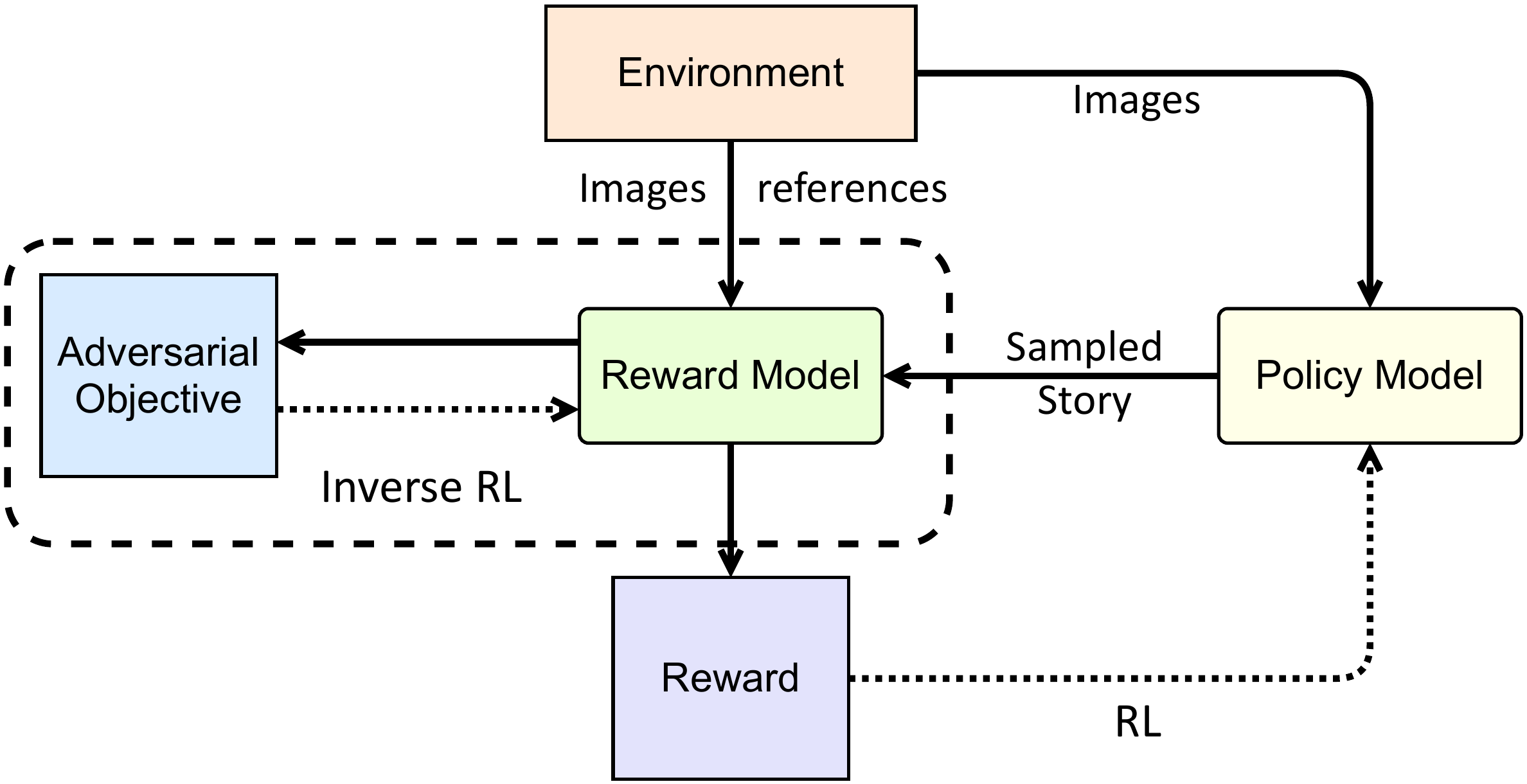}  
\end{center}
\vspace*{-1ex}
   \caption{AREL framework for visual storytelling.}
\label{fig:AREL}
\end{figure}

\section{Our Approach}
\subsection{Problem Statement}
Here we consider the task of visual storytelling, whose objective is to output a word sequence $W=(w_1, w_1, \cdots, w_T)$, $w_t \in \mathbb{V}$ given an input image stream of 5 ordered images $I=(I_1, I_2, \cdots, I_5)$, where $\mathbb{V}$ is the vocabulary of all output token. We formulate the generation as a markov decision process and design a reinforcement learning framework to tackle it. As described in~\autoref{fig:AREL}, our AREL framework is mainly composed of two modules: a \textbf{policy model} $\pi_{\beta}(W)$ and a \textbf{reward model} $R_{\theta}(W)$. The policy model takes an image sequence $I$ as the input and performs sequential actions (choosing words $w$ from the vocabulary $\mathbb{V}$) to form a narrative story $W$. The reward model is optimized by the adversarial objective (see Section~\ref{sec:learn}) and aims at deriving a human-like reward from both human-annotated stories and sampled predictions.
\begin{figure}[!t]
\begin{center}
\includegraphics[width=1.0\linewidth]{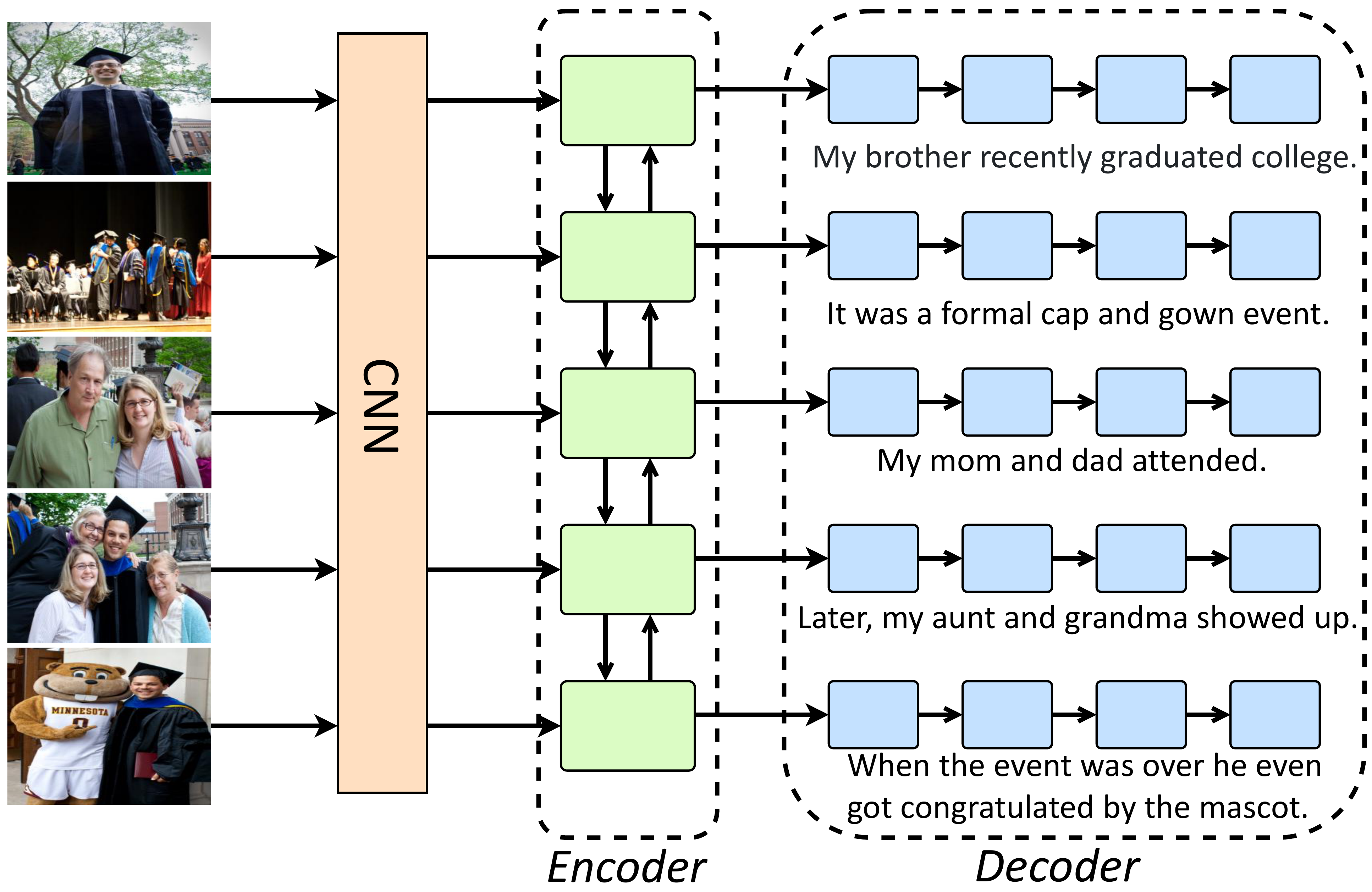}  
\end{center}
   \caption{Overview of the policy model. The visual encoder is a bidirectional GRU, which encodes the high-level visual features extracted from the input images. Its outputs are then fed into the RNN decoders to generate sentences in parallel. Finally, we concatenate all the generated sentences as a full story. Note that the five decoders share the same weights.}
\label{fig:writer}
\end{figure}

\subsection{Model}
\paragraph{Policy Model} As is shown in \autoref{fig:writer}, the policy model is a CNN-RNN architecture. We fist feed the photo stream $I = (I_1, \cdots, I_5)$ into a pretrained CNN and extract their high-level image features. We then employ a visual encoder to further encode the image features as context vectors $h_i=[\overleftarrow{h_i};\overrightarrow{h_i}]$. The visual encoder is a bidirectional gated recurrent units (GRU). 

In the decoding stage, we feed each context vector $h_i$ into a GRU-RNN decoder to generate a sub-story $W_i$. 
Formally, the generation process can be written as:
\begin{align}
s^i_t &= \text{GRU}(s^i_{t-1}, [w^i_{t-1}, h_i]) ~,    \\
\pi_{\beta}(w^i_{t}|w^i_{1:t-1}) &= softmax(W_s s^i_t + b_s) ~,
\end{align}
where $s^i_t$ denotes the $t$-th hidden state of $i$-th decoder. We concatenate the previous token $w^i_{t-1}$ and the context vector $h_i$ as the input. $W_s$ and $b_s$ are the projection matrix and bias, which output a probability distribution over the whole vocabulary $\mathbb{V}$. Eventually, the final story $W$ is the concatenation of the sub-stories $W_i$. $\beta$ denotes all the parameters of the encoder, the decoder, and the output layer.

\begin{figure}[!t]
\begin{center}
\includegraphics[width=1.0\linewidth]{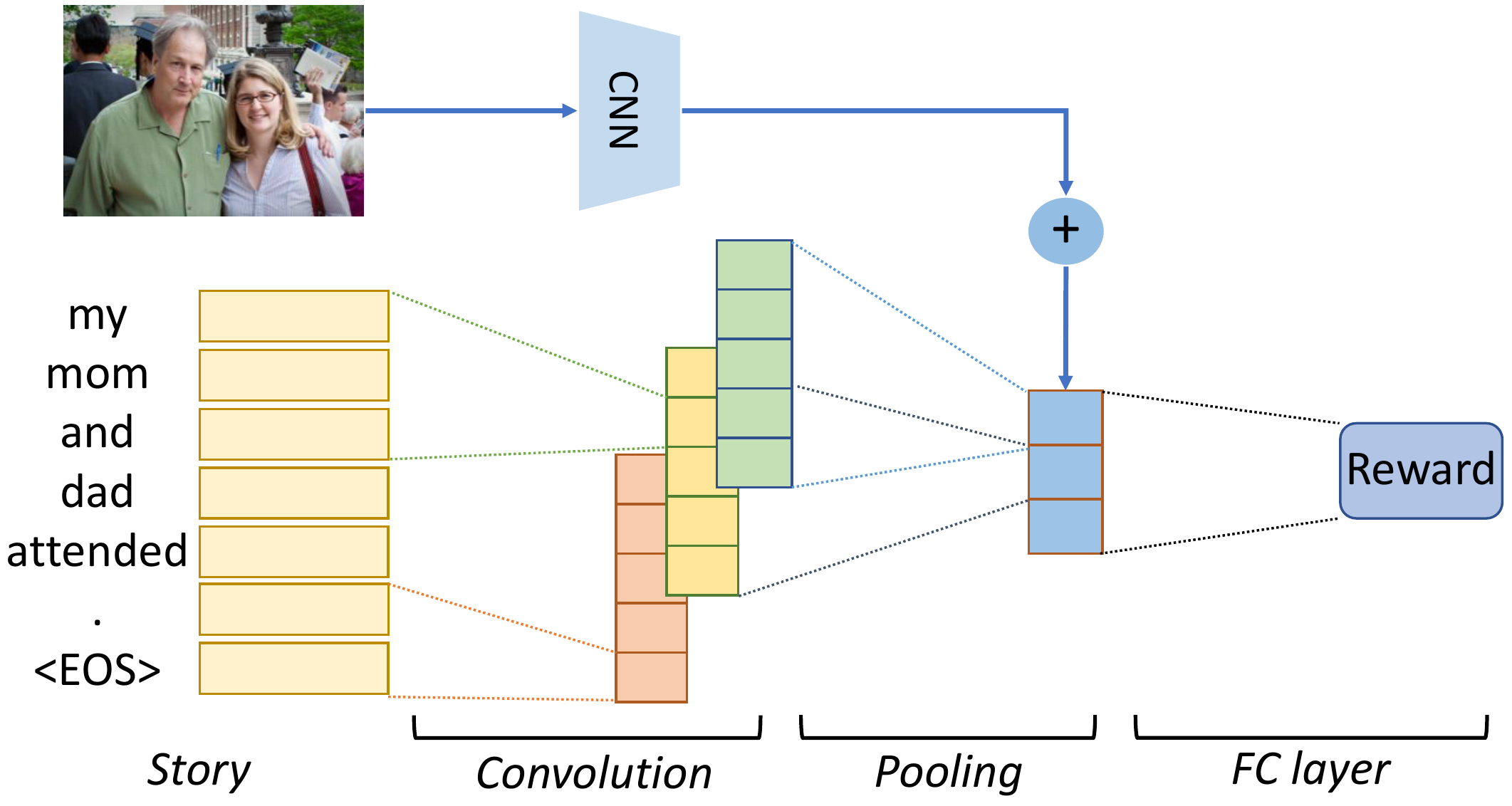}  
\end{center}
   \caption{Overview of the reward model. Our reward model is a CNN-based architecture, which utilizes convolution kernels with size 2, 3 and 4 to extract bigram, trigram and 4-gram representations from the input sequence embeddings. Once the sentence representation is learned, it will be concatenated with the visual representation of the input image, and then be fed into the final FC layer to obtain the reward.}
\label{fig:reviewer}
\end{figure}

\paragraph{Reward Model} The reward model $R_{\theta}(W)$ is a CNN-based architecture (see \autoref{fig:reviewer}).  
Instead of giving an overall score for the whole story, we apply the reward model to different story parts (sub-stories) $W_i$ and compute partial rewards, where $i = 1,\cdots, 5$. We observe that the partial rewards are more fine-grained and can provide better guidance for the policy model. 

We first query the word embeddings of the sub-story (one sentence in most cases). Next, multiple convolutional layers with different kernel sizes are used to extract the n-grams features, which are then projected into the sentence-level representation space by pooling layers (the design here is inspired by~\citet{kim2014convolutional}). In addition to the textual features, evaluating the quality of a story should also consider the image features for relevance. Therefore, we then combine the sentence representation with the visual feature of the input image through concatenation and feed them into the final fully connected decision layer. In the end, the reward model outputs an estimated reward value $R_{\theta}(W)$. The process can be written in formula:
\begin{align}
\small
\begin{split}
R_{\theta}(W) = \phi(W_r (f_{conv}(W) + W_i I_{CNN}) + b_r),
\end{split}
\end{align}
where $\phi$ denotes the non-linear projection function, $W_r, b_r$ denote the weight and bias in the output layer, and $f_{conv}$ denotes the operations in CNN. $I_{CNN}$ is the high-level visual feature extracted from the image, and $W_i$ projects it into the sentence representation space. $\theta$ includes all the parameters above.
\subsection{Learning}
\label{sec:learn}
\paragraph{Reward Boltzmann Distribution}
In order to associate story distribution with reward function, we apply EBM to define a Reward Boltzmann distribution:
\begin{align}
p_{\theta}(W) = \frac{\exp(R_{\theta}(W))}{Z_{\theta}} ~,
\end{align}
Where $W$ is the word sequence of the story and $p_{\theta}(W)$ is the approximate data distribution, and $Z_{\theta}=\underset{W}{\sum} \exp(R_{\theta}(W))$ denotes the partition function. According to the energy-based model~\cite{lecun2006tutorial}, the optimal reward function $R^*(W)$ is achieved when the Reward-Boltzmann distribution equals to the ``real'' data distribution $p_{\theta}(W) = p^*(W)$. 

\paragraph{Adversarial Reward Learning}
We first introduce an empirical distribution $p_e(W) = \frac{\mathbbm{1}(W \in D)}{|D|}$ to represent the empirical distribution of the training data, where $D$ denotes the dataset with $|D|$ stories and $\mathbbm{1}$ denotes an indicator function. We use this empirical distribution as the ``good'' examples, which provides the evidence for the reward function to learn from. 

In order to approximate the Reward Boltzmann distribution towards the ``real'' data distribution $p^*(W)$, we design a min-max two-player game, where the Reward Boltzmann distribution $p_{\theta}$ aims at maximizing the its similarity with empirical distribution $p_e$ while minimizing that with the ``faked" data generated from policy model $\pi_{\beta}$. On the contrary, the policy distribution $\pi_{\beta}$ tries to maximize its similarity with the Boltzmann distribution $p_{\theta}$. Formally, the adversarial objective function is defined as
\vspace*{-1ex}
\begin{align}
\small
\begin{split}
\max_{\beta} \min_{\theta} KL(p_e(W)||p_{\theta}(W)) - KL(\pi_{\beta}(W)||p_{\theta}(W)) ~.
\end{split}
\end{align}
We further decompose it into two parts. First, because the objective $J_{\beta}$ of the story generation policy is to maximize its similarity with the Boltzmann distribution $p_{\theta}$, the optimal policy that minimizes KL-divergence is thus $\pi(W) \sim \exp(R_{\theta}(W))$, meaning if $R_{\theta}$ is optimal, the optimal $\pi_{\beta} = \pi^*$. In formula, 
\begin{align} 
\small
\begin{split}
J_{\beta}
=& -KL(\pi_{\beta}(W)||p_{\theta}(W)) \\
=& \expect{W \sim \pi_{\beta}(W)} [R_{\theta}(W)] - \log Z_\theta + H(\pi_{\beta}(W)) ~,
\end{split}
\end{align}
where $H$ denotes the entropy of the policy model. On the other hand, the objective $J_{\theta}$ of the reward function is to distinguish between human-annotated stories and machine-generated stories. Hence it is trying to minimize the KL-divergence with the empirical distribution $p_e$ and maximize the KL-divergence with the approximated policy distribution $\pi_{\beta}$:
\begin{align} 
\small
\begin{split}
J_{\theta}
=& KL(p_e(W)||p_{\theta}(W)) - KL(\pi_{\beta}(W)||p_{\theta}(W)) \\
=&\sum_{W} [p_e(W) R_{\theta}(W) - \pi_{\beta}(W) R_{\theta}(W)] \\
 & + \log Z_\theta - \log Z_\theta - H(p_e) + H(\pi_{\beta}) ~,
\end{split}
\end{align}
Since $H(\pi_{\beta})$ and $H(p_e)$ are irrelevant to $\theta$, we denote them as constant $C$. It is also worth noting that with negative sampling in the optimization of the KL-divergence, the computation of the intractable partition function $Z_{\theta}$ is bypassed. Therefore, the objective $J_{\theta}$ can be further derived as
\begin{align} 
\small
\begin{split}
J_{\theta}
=& \expect{W \sim p_e(W)}[R_{\theta}(W)] - \expect{W \sim \pi_{\beta}(W)}[R_{\theta}(W)] + C ~.
\end{split}
\end{align}

Here we propose to use stochastic gradient descent to optimize these two models alternately. Formally, the gradients can be written as
\begin{align}
\label{eq:gradient}
\small
\begin{split}
\frac{\partial J_{\theta}}{\partial \theta} =& \expect{W \sim p_e(W)}[\frac{\partial R_{\theta}(W)}{\partial \theta}] - \expect{W \sim \pi_{\beta}(W)}[\frac{\partial R_{\theta}(W)}{\partial \theta}] ~,
\\ 
\frac{\partial J_{\beta}}{\partial \beta} =& \expect{W \sim \pi_{\beta}(W)} (R_{\theta}(W) - \log \pi_{\beta}(W) - b) \frac{\partial \log \pi_{\beta}(W)}{\partial \beta} ~,
\end{split}
\end{align}
where $b$ is the estimated baseline to reduce variance during REINFORCE training.

\begin{algorithm}[t]
\begin{algorithmic}[1]
\For{episode $\leftarrow$ 1 to N}
\State collect story $W$ by executing policy $\pi_{\theta}$
\If{Train-Reward}
\State $\theta \leftarrow \theta - \eta  \times \frac{\partial J_{\theta}}{\partial \theta}$  (see \autoref{eq:gradient})
\ElsIf{Train-Policy}
\State collect story $\tilde{W}$ from empirical $p_e$
\State $\beta \leftarrow \beta - \eta \times \frac{\partial J_{\beta}}{\partial \beta}$ (see \autoref{eq:gradient}) 
\EndIf
\EndFor
\end{algorithmic}
\caption{The AREL Algorithm.}\label{alg:airl}
\end{algorithm}

\paragraph{Training \& Testing} As described in \autoref{alg:airl}, we introduce an alternating algorithm to train these two models using stochastic gradient descent. During testing, the policy model is used with beam search to produce the story.

\section{Experiments and Analysis}

\label{sec:exp}
\subsection{Experimental Setup}
\paragraph{VIST Dataset} The VIST dataset~\cite{huang2016visual} is the first dataset for sequential vision-to-language tasks including visual storytelling, which consists of 10,117 Flickr albums with 210,819 unique photos. In this paper, we mainly evaluate our AREL method on this dataset. After filtering the broken images\footnote{There are only 3 (out of 21,075) broken images in the test set, which basically has no influence on the final results. Moreover, \citet{yu-bansal-berg:2017:EMNLP2017} also removed the 3 pictures, so it is a fair comparison.}, there are 40,098 training, 4,988 validation, and 5,050 testing samples. 
Each sample contains one story that describes 5 selected images from a photo album (mostly one sentence per image). And the same album is paired with 5 different stories as references. In our experiments, we used the same split settings as in~\cite{huang2016visual,yu-bansal-berg:2017:EMNLP2017} for a fair comparison. During our experiments, we apply two kinds of non-linear functions $\phi$ for the discriminator, namely SoftSign function ($f(x)=\frac{x}{1+|x|}$) and Hyperbolic function  ($f(x)=\frac{sinh x}{cosh x}$). We found that unbounded non-linear functions like ReLU function~\cite{glorot2011deep} will lead to severe vibrations and instabilities during training, therefore we resort to the bounded functions.

\paragraph{Evaluation Metrics}
In order to comprehensively evaluate our method on storytelling dataset, we adopt both the automatic metrics and human evaluation as our criterion. Four diverse automatic metrics are used in our experiments: BLEU, METEOR, ROUGE-L, and CIDEr. We utilize the open source evaluation code\footnote{\scriptsize\url{https://github.com/lichengunc/vist_eval}} used in~\cite{yu-bansal-berg:2017:EMNLP2017}. For human evaluation, we employ the Amazon Mechanical Turk to perform two kinds of user studies (see Section~\ref{sec:human} for more details). 

\paragraph{Training Details}
We employ pretrained ResNet-152 model~\cite{he2016deep} to extract image features from the photostream. We built a vocabulary of size 9,837 to include words appearing more than three times in the training set. More training details can be found at \autoref{supp:training}. 

\subsection{Automatic Evaluation}
In this section, we compare our AREL method with the state-of-the-art methods as well as standard reinforcement learning algorithms on automatic evaluation metrics. Then we further discuss the limitations of the hand-crafted metrics on evaluating human-like stories.

\paragraph{Comparison with SOTA on Automatic Metrics}
In~\autoref{table:sota}, we compare our method with \citet{huang2016visual} and \citet{yu-bansal-berg:2017:EMNLP2017}, which report achieving best-known results on the VIST dataset. We first implement a strong baseline model (\textit{XE-ss}), which share the same architecture with our policy model but is trained with cross-entropy loss and scheduled sampling. Besides, we adopt the traditional generative adversarial training for comparison (\textit{GAN}). As shown in~\autoref{table:sota}, our XE-ss model already outperforms the best-known results on the VIST dataset, and the GAN model can bring a performance boost. We then use the XE-ss model to initialize our policy model and further train it with \textit{AREL}. Evidently, our AREL model performs the best and achieves the new state-of-the-art results across all metrics. 

\begin{table} 
\small
\renewcommand{\arraystretch}{1.1}
\begin{center}
  \begin{tabular}{ l @{\hspace{0.25cm}} | c @{\hspace{0.25cm}} c @{\hspace{0.25cm}} c @{\hspace{0.25cm}} c @{\hspace{0.25cm}} c @{\hspace{0.25cm}} c @{\hspace{0.25cm}} c }
  
   Method     & B-1 & B-2 & B-3 & B-4 & M & R & C \\
   \hline\hline
   Huang et al.
   & - & - & - & - & 31.4 & - & -  \\
   Yu et al.       
   & - & - & 21.0 & - & 34.1 & 29.5 & 7.5  \\
   \hline
   XE-ss     
   & 62.3 & 38.2 & 22.5 & 13.7 & 34.8 & \textbf{29.7} & 8.7 \\
   GAN
   & 62.8 & 38.8 & 23.0 & 14.0 & 35.0 & 29.5 & 9.0 \\
   AREL-s-50 
   & 62.9 & 38.4 & 22.7 & 14.0 & 34.9 & 29.4 & 9.1 \\
   AREL-t-50
   & 63.4 & 39.0 & 23.1 & \textbf{14.1} & \textbf{35.2} & 29.6 & 9.5 \\
   AREL-s-100
   & \textbf{64.0} & 38.6 & 22.3 & 13.2 & 35.1 & 29.3 & \textbf{9.6} \\
   AREL-t-100
   & 63.8 & \textbf{39.1} & \textbf{23.2} & \textbf{14.1} & 35.0 & 29.5 & 9.4 \\ 
  \end{tabular}
\end{center}
\caption{Automatic evaluation on the VIST dataset. We report BLEU (B), METEOR (M), ROUGH-L (R), and CIDEr (C) scores of the SOTA systems and the models we implemented, including XE-ss, GAN and AREL. AREL-s-N denotes AREL models with SoftSign as output activation and alternate frequency as N, while AREL-t-N denoting AREL models with Hyperbolic as the output activation (N = 50 or 100).}
\label{table:sota}
\end{table}

But, compared with the XE-ss model, the performance gain is minor, especially on METEOR and ROUGE-L scores. However, in Sec.~\ref{sec:human}, the extensive human evaluation has indicated that our AREL framework brings a significant improvement on generating human-like stories over the XE-ss model. The inconsistency of automatic evaluation and human evaluation lead to a suspect that these hand-crafted metrics lack the ability to fully evaluate stories' quality due to the complicated characteristics of the stories. Therefore, we conduct experiments to analyze and discuss the defects of the automatic metrics in~\autoref{sec:limit}.

\begin{table} 
\small
\renewcommand{\arraystretch}{1.1}
\begin{center}
 \begin{tabular}{ l @{\hspace{0.25cm}} | c @{\hspace{0.25cm}} c @{\hspace{0.25cm}} c @{\hspace{0.25cm}} c @{\hspace{0.25cm}} c @{\hspace{0.25cm}} c @{\hspace{0.25cm}} c }
  
  Method         & B-1 & B-2 & B-3 & B-4 & M & R & C \\
  \hline\hline
  XE-ss     & 62.3 & 38.2 & 22.5 & 13.7 & 34.8 & \textbf{29.7} & 8.7 \\
  \hline
  BLEU-RL      & 62.1 & 38.0 & 22.6 & 13.9 & 34.6 & 29.0 & 8.9 \\
   METEOR-RL       & \textbf{68.1} & 35.0 & \focus{15.4} & \focus{6.8} & \textbf{40.2} & 30.0 & \focus{1.2} \\ 
   ROUGE-RL        & 58.1 & \focus{18.5} & \focus{1.6} & \focus{0} & \focus{27.0} & \textbf{33.8} & \focus{0} \\ 
   CIDEr-RL        & 61.9 & 37.8 & 22.5 & 13.8 & 34.9 & 29.7 & 8.1 \\
   \hline 
   AREL (best)      & \textbf{63.8} & \textbf{39.1} & \textbf{23.2} & \textbf{14.1} & \textbf{35.0} & 29.5 & \textbf{9.4} \\ 
 \end{tabular}
\end{center}
\caption{Comparison with different RL models with different metric scores as the rewards. We report the average scores of the AREL models as AREL (avg). Although METEOR-RL and ROUGE-RL models achieve the highest scores on their own metrics, the underlined scores are severely damaged. Actually, they are gaming their own metrics with nonsense sentences.}
\label{table:rl}
\end{table}

\paragraph{Limitations of Automatic Metrics}
\label{sec:limit}
String-match-based automatic metrics are not perfect and fail to evaluate some semantic characteristics of the stories (e.g. expressiveness and coherence). In order to confirm our conjecture, we utilize automatic metrics as rewards to reinforce the model with policy gradient. The quantitative results are demonstrated in \autoref{table:sota}. 

Apparently, METEOR-RL and ROUGE-RL are severely ill-posed: they obtain the highest scores on their own metrics but damage the other metrics severely. We observe that these models are actually overfitting to a given metric while losing the overall coherence and semantical correctness. Same as METEOR score, there is also an adversarial example for ROUGE-L\footnote{An adversarial example for ROUGE-L: \textit{we the was a . and to the . we the was a . and to the . we the was a . and to the . we the was a . and to the . we the was a . and to the .}}, which is nonsense but achieves an average ROUGE-L score of 33.8.

\begin{figure*}
\vspace{-1ex}
\centering
\includegraphics[width=1\linewidth]{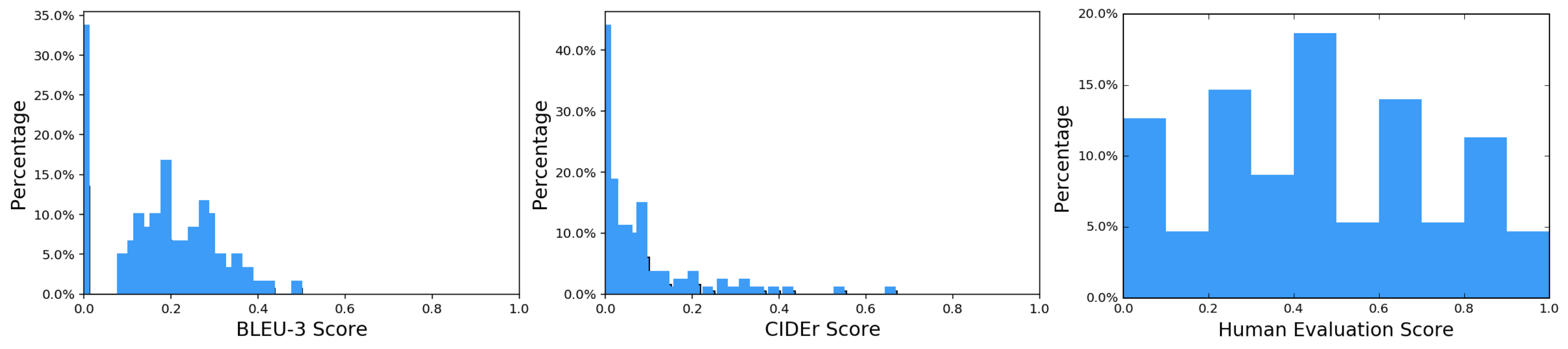}  
\vspace*{-4ex}
\caption{Metric score distributions. We plot the histogram distributions of BLEU-3 and CIDEr scores on the test set, as well as the human evaluation score distribution on the test samples. We use the Turing test results to calculate the human evaluation scores (see Section~\ref{sec:human}). Basically, 0.2 score is given if the generated story wins the Turing test, 0.1 for tie, and 0 if losing. Each sample has 5 scores from 5 judges, and we use the sum as the human evaluation score, so it is in the range [0, 1]. }
\label{fig:score}
\end{figure*}

Besides, as can be seen in \autoref{table:sota}, after reinforced training, BLEU-RL and CIDEr-RL do not bring a consistent improvement over the XE-ss model. We plot the histogram distributions of both BLEU-3 and CIDEr scores on the test set in \autoref{fig:score}. 
An interesting fact is that there are a large number of samples with nearly zero score on both metrics. However, we observed those ``zero-score" samples are not pointless results; instead, lots of them make sense and deserve a better score than zero. Here is a ``zero-score" example on BLEU-3:
\begin{myquote}{0.2in}
\textit{I had a great time at the restaurant today. The food was delicious. I had a lot of food. The food was delicious. I had a great time.} 
\end{myquote}
The corresponding reference is 
\begin{myquote}{0.2in}
\textit{The table of food was a pleasure to see! Our food is both nutritious and beautiful! Our chicken was especially tasty! We love greens as they taste great and are healthy! The fruit was a colorful display that tantalized our palette.}
\end{myquote}
Although the prediction is not as good as the reference, it is actually coherent and relevant to the theme ``food and eating", which showcases the defeats of using BLEU and CIDEr scores as a reward for RL training.

\begin{table} 
\small
\begin{center}
  \begin{tabular}{ l | c  c  c}
  
    Method     & Win & Lose & Unsure \\
    \hline
    XE-ss     & 22.4\% & 71.7\% & 5.9\% \\
    BLEU-RL    & 23.4\% & 67.9\% & 8.7\% \\
    CIDEr-RL& 13.8\% & 80.3\% & 5.9\% \\
    GAN        & 34.3\% & 60.5\% & 5.2\% \\
    \hline
    AREL     & \textbf{38.4}\% & \textbf{54.2}\% & \textbf{7.4}\% \\ 
    
  \end{tabular}
\end{center}
\caption{Turing test results.}
\label{table:turing}
\end{table}

\begin{table*}
\small
\renewcommand{\arraystretch}{1.1}
\centering

\begin{tabular}{ l @{\hspace{0.25cm}} | c @{\hspace{0.25cm}} c @{\hspace{0.25cm}} c | c @{\hspace{0.25cm}} c  @{\hspace{0.25cm}} c | c @{\hspace{0.25cm}} c @{\hspace{0.25cm}} c | c @{\hspace{0.25cm}} c @{\hspace{0.25cm}} c }

               & \multicolumn{3}{c|}{AREL \textit{vs} XE-ss} & \multicolumn{3}{c|}{AREL \textit{vs} BLEU-RL} & \multicolumn{3}{c|}{AREL \textit{vs} CIDEr-RL} & \multicolumn{3}{c}{AREL \textit{vs} GAN} \\
           \hline
           Choice (\%)　& AREL & XE-ss & Tie & AREL & BLEU-RL & Tie & AREL &CIDEr-RL & Tie  & AREL &GAN & Tie \\
          \hline
         Relevance         
         & \textbf{61.7}     & 25.1      & 13.2
         & \textbf{55.8}     & 27.9      & 16.3
         & \textbf{56.1}     & 28.2      & 15.7
         & \textbf{52.9}     & 35.8      & 11.3 \\ 
         Expressiveness    
         & \textbf{66.1}     & 18.8      & 15.1
         & \textbf{59.1}     & 26.4      & 14.5
         & \textbf{59.1}     & 26.6      & 14.3
         & \textbf{48.5}     & 32.2      & 19.3 \\
         Concreteness      
         & \textbf{63.9}     & 20.3     & 15.8 
         & \textbf{60.1}     & 26.3      & 13.6
         & \textbf{59.5}     & 24.6      & 15.9
         & \textbf{49.8}     & 35.8      & 14.4 \\

\end{tabular}
\caption{Pairwise human comparisons. The results indicate the consistent superiority of our AREL model in generating more human-like stories than the SOTA methods.}
\label{table:human}
\end{table*}

Moreover, we compare the human evaluation scores with these two metric scores in \autoref{fig:score}. Noticeably, both BLEU-3 and CIDEr have a poor correlation with the human evaluation scores. Their distributions are more biased and thus cannot fully reflect the quality of the generated stories. In terms of BLEU, it is extremely hard for machines to produce the exact 3-gram or 4-gram matching, so the scores are too low to provide useful guidance. 
CIDEr measures the similarity of a sentence to the majority of the references. 
However, the references to the same image sequence are photostream different from each other, so the score is very low and not suitable for this task. In contrast, our AREL framework can lean a more robust reward function from human-annotated stories, which is able to provide better guidance to the policy and thus improves its performances over different metrics. 

\begin{figure}
\centering
\includegraphics[width=0.9\linewidth]{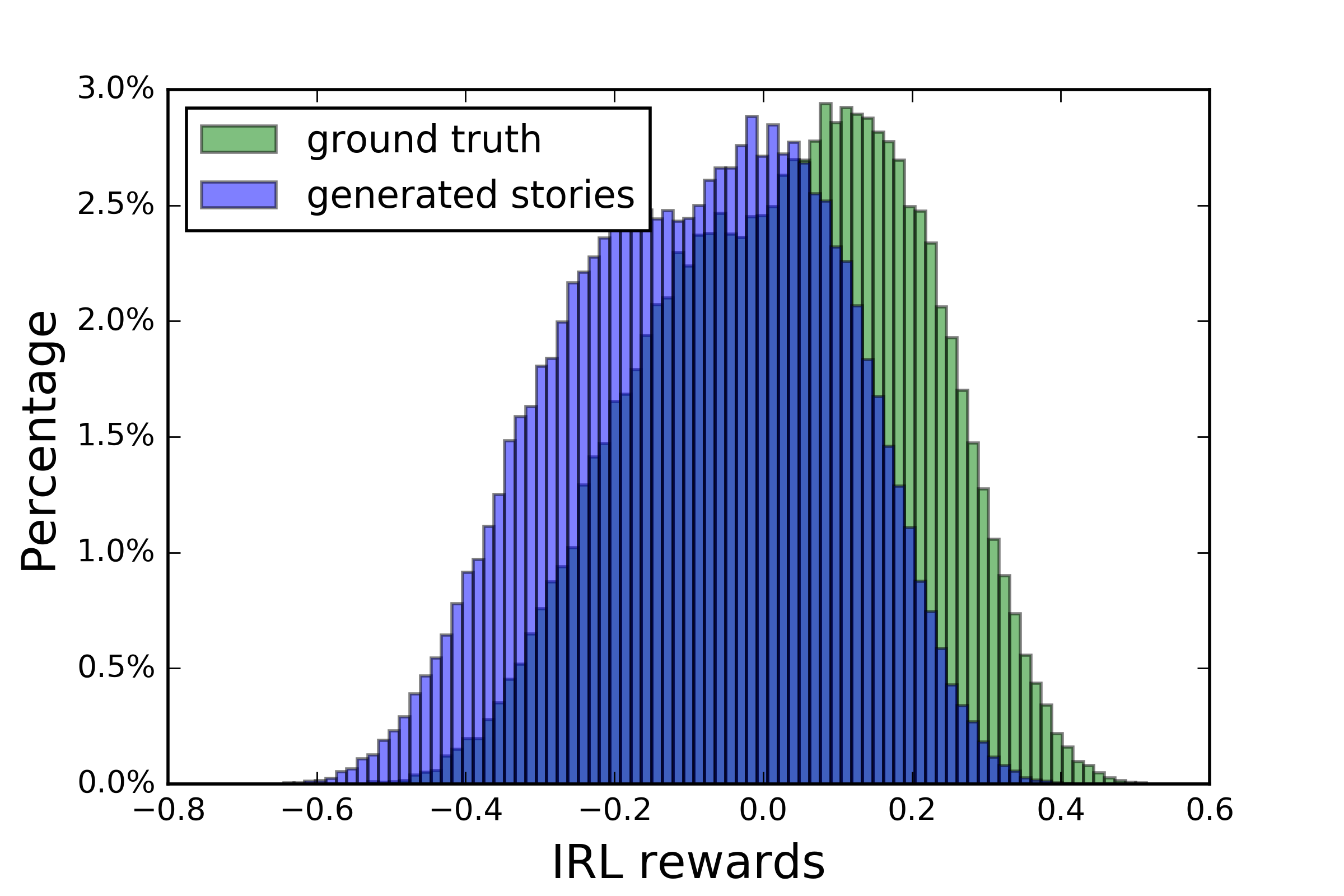}  
\vspace*{-1ex}
\caption{Visualization of the learned rewards on both the ground-truth stories and the stories generated by our AREL model. The generated stories are receiving lower averaged scores than the human-annotated ones.}
\label{fig:visualize-reward}
\end{figure}

\paragraph{Visualization of The Learned Rewards}
In~\autoref{fig:visualize-reward}, we visualize the learned reward function for both ground truth and generated stories. Evidently, the AREL model is able to learn a smoother reward function that can distinguish the generated stories from human annotations. In other words, the learned reward function is more in line with human perception and thus can encourage the model to explore more diverse language styles and expressions.   

\paragraph{Comparison with GAN}
We here compare our method with vanilla GAN~\cite{goodfellow2014generative}, whose update rules for the generator can be generally classified into two categories. We demonstrate their corresponding objectives and ours as follows:
\begin{align*}
\small
GAN1: \quad J_{\beta} =& \expect{W \sim p_{\beta}}[-\log R_{\theta}(W)] ~, \\
GAN2: \quad J_{\beta} =& \expect{W \sim p_{\beta}}[\log (1 - R_{\theta}(W))] ~, \\
ours: \quad J_{\beta} =& \expect{W \sim p_{\beta}}[- R_{\theta}(W)] ~.
\end{align*}
As discussed in \citet{arjovsky2017wasserstein}, $GAN1$ is prone to the unstable gradient issue and $GAN2$ is prone to the vanishing gradient issue. Analytically, our method does not suffer from these two common issues and thus is able converge to optimum solutions more easily. From \autoref{table:sota} we can observe slight gains of using AREL over GAN with automatic metrics, but we further deploy human evaluation for a better comparison. 

\subsection{Human Evaluation}
\label{sec:human}
Automatic metrics cannot fully evaluate the capability of our AREL method. Therefore, we perform two different kinds of human evaluation studies on Amazon Mechanical Turk: Turing test and pairwise human evaluation. For both tasks, we use 150 stories (750 images) sampled from the test set, each assigned to 5 workers to eliminate human variance. We batch six items as one assignment and insert an additional assignment as a sanity check. Besides, the order of the options within each item is shuffled to make a fair comparison.

\begin{figure*}
\centering
  \includegraphics[width=1\textwidth]{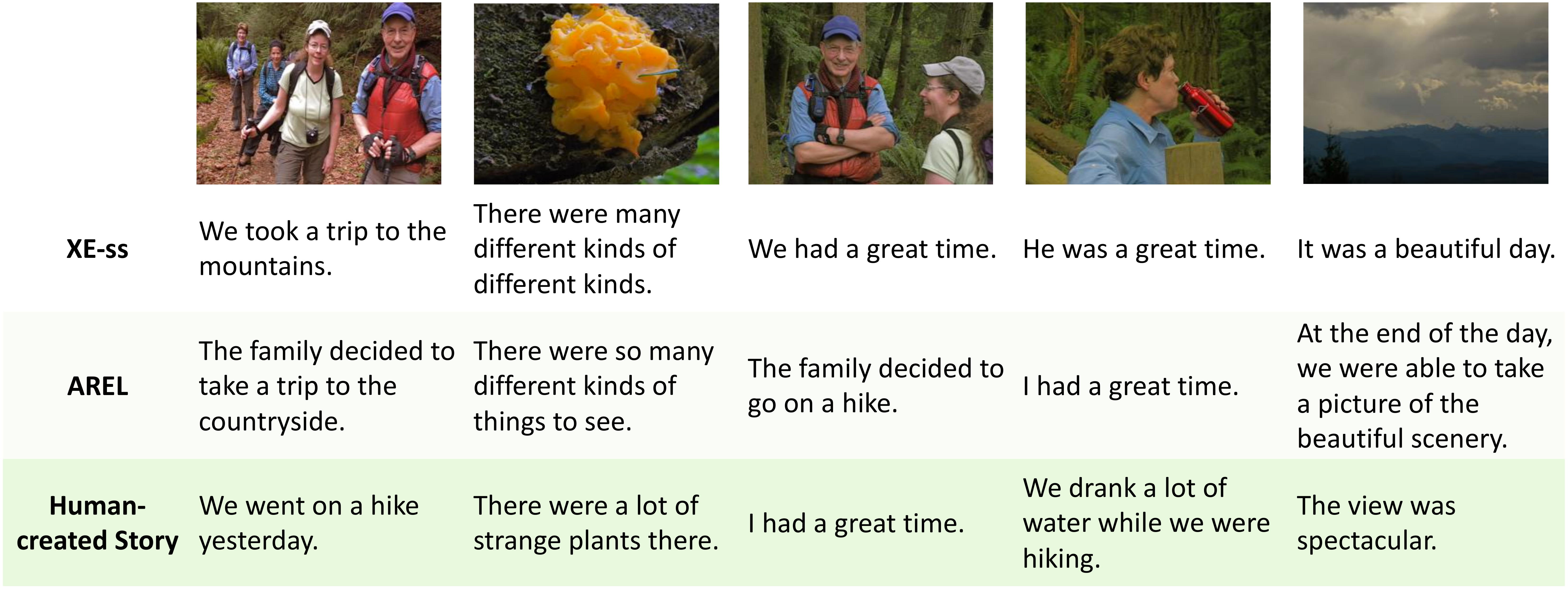}  
  \vspace{-1ex}
  \caption{Qualitative comparison example with XE-ss. The direct comparison votes (AREL:XE-ss:Tie) were 5:0:0 on Relevance, 4:0:1 on Expressiveness, and 5:0:0 on Concreteness. }
  \label{fig:pos}
\end{figure*}

\paragraph{Turing Test} We first conduct five independent Turing tests for XE-ss, BLEU-RL, CIDEr-RL, GAN, and AREL models, during which the worker is given one human-annotated sample and one machine-generated sample, and needs to decide which is human-annotated. As shown in \autoref{table:turing}, our AREL model significantly outperforms all the other baseline models in the Turing test: it has much more chances to fool AMT worker (the ratio is AREL:XE-ss:BLEU-RL:CIDEr-RL:GAN = 45.8\%:28.3\%:32.1\%:19.7\%:39.5\%), which confirms the superiority of our AREL framework in generating human-like stories. Unlike automatic metric evaluation, the Turing test has indicated a much larger margin between AREL and other competing algorithms. Thus, we empirically confirm that metrics are not perfect in evaluating many implicit semantic properties of natural language.
Besides, the Turing test of our AREL model reveals that nearly half of the workers are fooled by our machine generation, indicating a preliminary success toward generating human-like stories. 

\paragraph{Pairwise Comparison} In order to have a clear comparison with competing algorithms with respect to different semantic features of the stories, we further perform four pairwise comparison tests: AREL \textit{vs} XE-ss/BLEU-RL/CIDEr-RL/GAN. For each photostream, the worker is presented with two generated stories and asked to make decisions from the  three aspects: relevance\footnote{Relevance: the story accurately describes what is happening in the image sequence and covers the main objects.}, expressiveness\footnote{Expressiveness: coherence, grammatically and semantically correct, no repetition, expressive language style.} and concreteness\footnote{Concreteness: the story should narrate concretely what is in the image rather than giving very general descriptions.}. This head-to-head compete is designed to help us understand in what aspect our model outperforms the competing algorithms, which is displayed in~\autoref{table:human}.

Consistently on all the three comparisons, a large majority of the AREL stories trumps the competing systems with respect to their relevance, expressiveness, and concreteness. Therefore, it empirically confirms that our generated stories are more relevant to the image sequences, more coherent and concrete than the other algorithms, which however is not explicitly reflected by the automatic metric evaluation.

\subsection{Qualitative Analysis}
\autoref{fig:pos} gives a qualitative comparison example between AREL and XE-ss models. Looking at the individual sentences, it is obvious that our results are more grammatically and semantically correct. Then connecting the sentences together, we observe that the AREL story is more coherent and describes the photo stream more accurately. Thus, our AREL model significantly surpasses the XE-ss model on all the three aspects of the qualitative example. Besides, it won the Turing test (3 out 5 AMT workers think the AREL story is created by a human). 
In the appendix, we also show a negative case that fails the Turing test. 

\section{Conclusion}
In this paper, we not only introduce a novel adversarial reward learning algorithm to generate more human-like stories given image sequences, but also empirically analyze the limitations of the automatic metrics for story evaluation. We believe there are still lots of improvement space in the narrative paragraph generation tasks, like how to better simulate human imagination to create more vivid and diversified stories.

\section*{Acknowledgment}
We thank Adobe Research for supporting our language and vision research. We would also like to thank Licheng Yu for clarifying the details of his paper and the anonymous reviewers for their thoughtful comments. This research was sponsored in part by the Army Research Laboratory under cooperative agreements W911NF09-2-0053. The views and conclusions contained herein are those of the authors and should not be interpreted as representing the official policies, either expressed or implied, of the Army Research Laboratory or the U.S. Government. The U.S. Government is authorized to reproduce and distribute reprints for Government purposes notwithstanding any copyright notice herein.

\bibliography{acl2018}
\bibliographystyle{acl_natbib}


\newpage
\clearpage
\appendix

\section*{Appendix}
\label{sec:supp}

\begin{figure*}
  \begin{center}
  \includegraphics[width=1\textwidth]{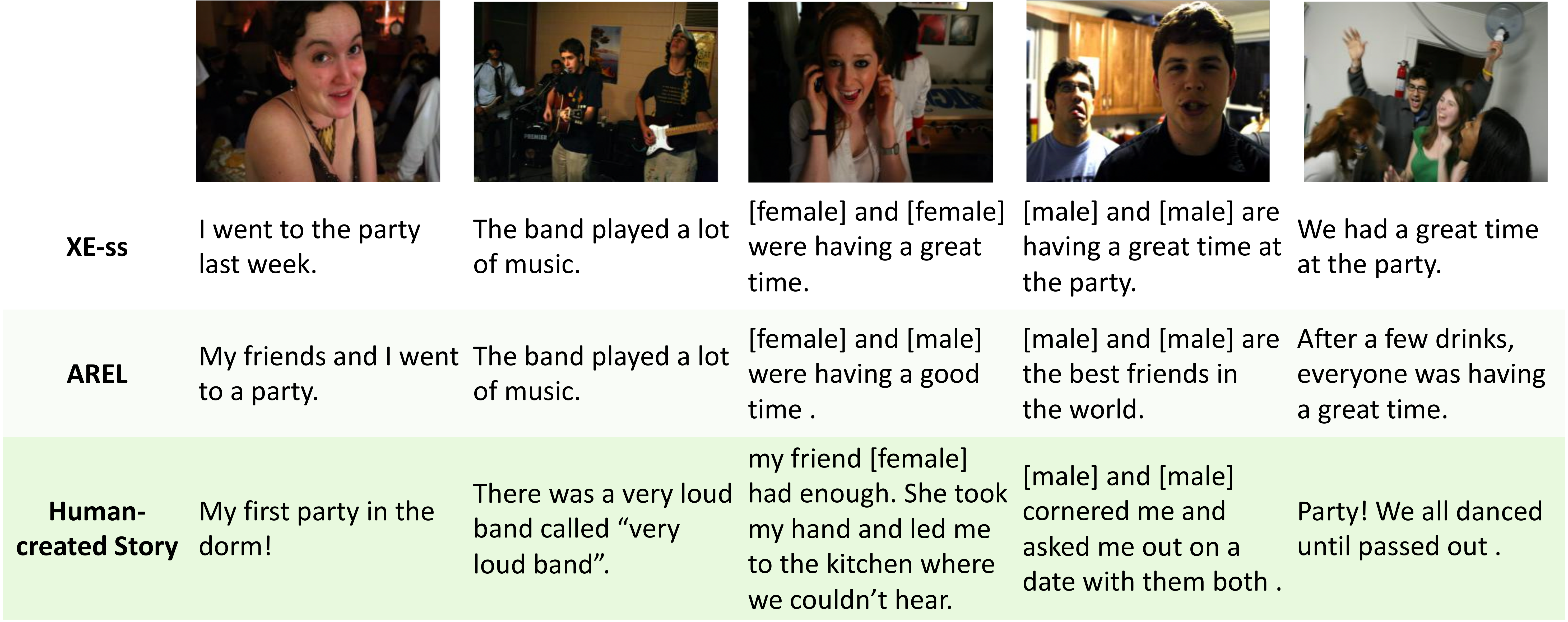}  
  \end{center}
     \caption{Failure case in Turing test. 4 out of 5 workers correctly recognized the human-created story and 1 person mistakenly chose AREL story. 
     }
  \label{fig:neg}
\end{figure*}

\section{Error Analysis}
\paragraph{Failure Case in Turing Test}
In \autoref{fig:neg}, we presented a negative example that failed the Turing test (4 out of 5 made the correct decision). Compared with the human-generated story, our AREL story lacked emotion and imagination and thus can be easily distinguished. For example, the real human gave the band a nickname ``very loud band" and told a more amusing story. Though we have made encouraging progress on generating human-like stories, further research of creating diversified stories is still needed.

\paragraph{Data Bias}
From the experiments, we observe that there exist some severe data bias issues in the VIST dataset, such as gender bias and event bias. In the training set, the ratio of male and female's appearances is 2.06:1, and it is 2.16:1 in the test set. the models aggravate the gender bias to 3.44:1. Besides, because all the images are collected from Flickr, there is also an event bias issue. We count three most frequent events: party, wedding, and graduation, whose ratios are 6.51:2.36:1 on the training set and 4.54:2.42:1 on the test set. However, their ratio on the testing results is 10.69:2.22:1. Clearly, the models tend to magnify the influence of the largest majority. These bias issues remain to be studied for future work.

\section{Training Details}
\label{supp:training}
Our model is implemented on PyTorch and consists of two parts -- a policy model and a reward model. The policy model is implemented with a multiple-RNN architecture. Each RNN model is responsible for generating a sub-story for each photo in the stream. But the weights are tied to minimize the memory consumption. The image features are extracted from the pre-trained ResNet-152 model\footnote{\scriptsize\url{https://github.com/KaimingHe/deep-residual-networks}}. The visual encoder receives the ResNet-152 features and uses recurrent neural network to understand the temporal dynamics and represents them as hidden state vectors, which is further fed into the decoder to generate stories. The reward model is based on convolutional neural network and uses convolution kernels to extract semantic features for prediction.
Here we give the detailed description of our system: 
\begin{itemize}
\item Visual Encoder: the visual encoder is a bi-directional GRU model with hidden dimension of 256 for each direction. we concatenate the bi-directional states and form a 512 dimension vector for the story generator. The input album is composed of five images, and each image is used as separate input to different RNN decoders.
\item Decoder: The decoder is a single-layer GRU model with hidden dimension of 512. The recurrent decoder model receives the output from the visual encoder as the first input, and then at the following time steps, it receives the last predicted token as input or uses the ground truth as input. During scheduled sampling, we use a sampling probability to decide which action to take.
\item Reward Model: we use a convolutional neural network to extract n-gram features from the story embedding and stretch them into a flattened vector. The embedding size of input story is 128, and the filter dimension of CNN is also 128. Here we use three kernels with window size 2, 3, 4, each with a stride size of 1. We use a pooling size of 2 to shrink the extracted outputs and flatten them as a vector. Finally, we project this vector into a single cell indicating the predicted reward value. 
\end{itemize}
During training, we first pre-train a schedule-sampling model with a batch size of 64 with NVIDIA Titan X GPU. The warm-up process takes roughly 5-10 hours, and then we select the best model to initialize our AREL policy model. Finally, we use alternating training strategy to optimize both the policy model and the reward model with a learning rate of 2e-4 using Adam optimization algorithm. During test time, we use a beam size of 3 to approximate the whole search space, we force the beam search to proceed more than 5 steps and no more than 110 steps. Once we reach the EOS token, the algorithm stops and we compare the results with human-annotated corpus using 4 different automatic evaluation metrics.

\section{Amazon Mechanical Turk}
\begin{figure*}
\centering
\includegraphics[width=1.0\linewidth]{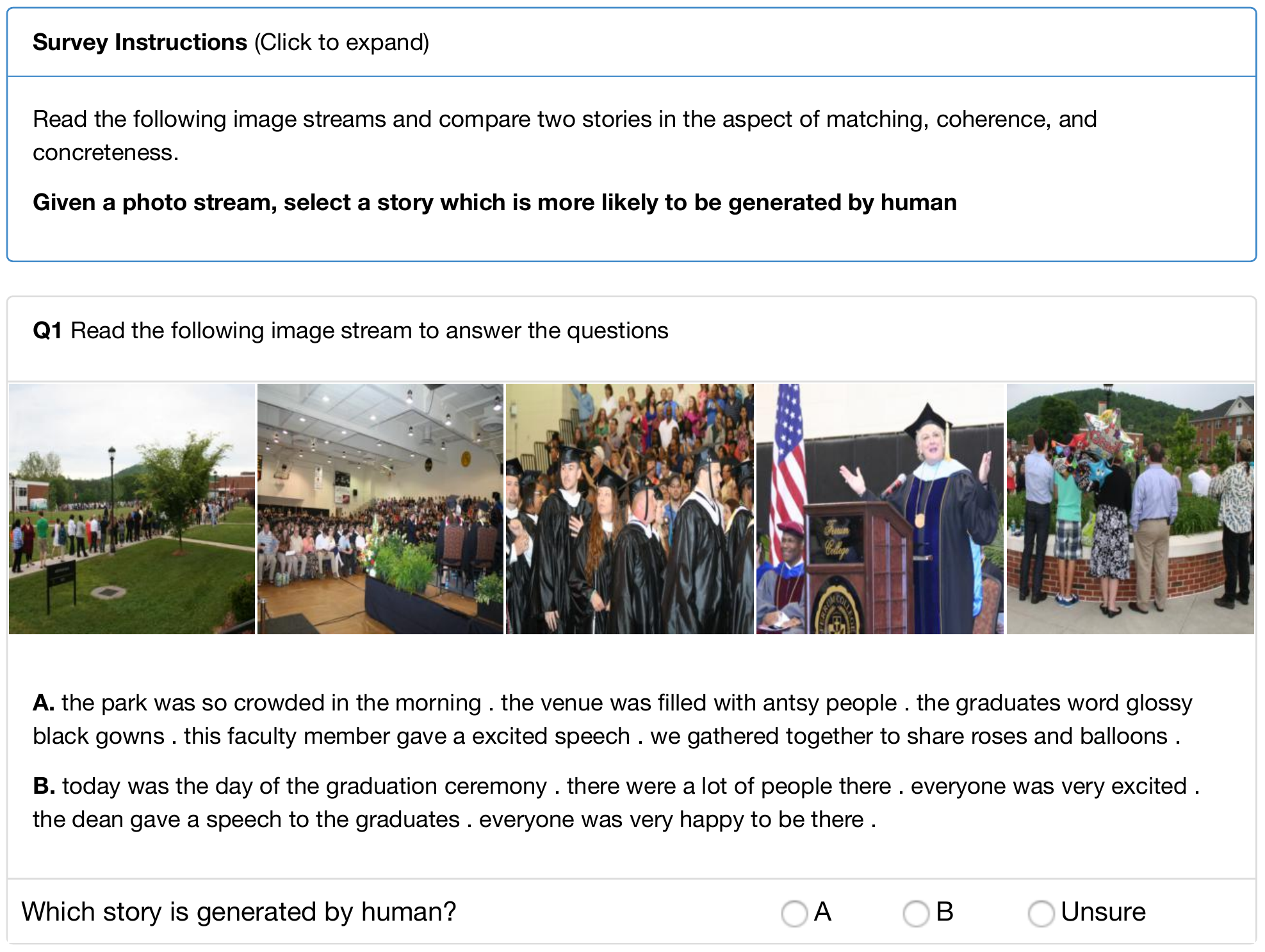}      
\caption{Turing Survey Form}
\label{fig:turing}
\end{figure*}

\begin{figure*}[!thb]
\centering
\includegraphics[width=1.0\linewidth]{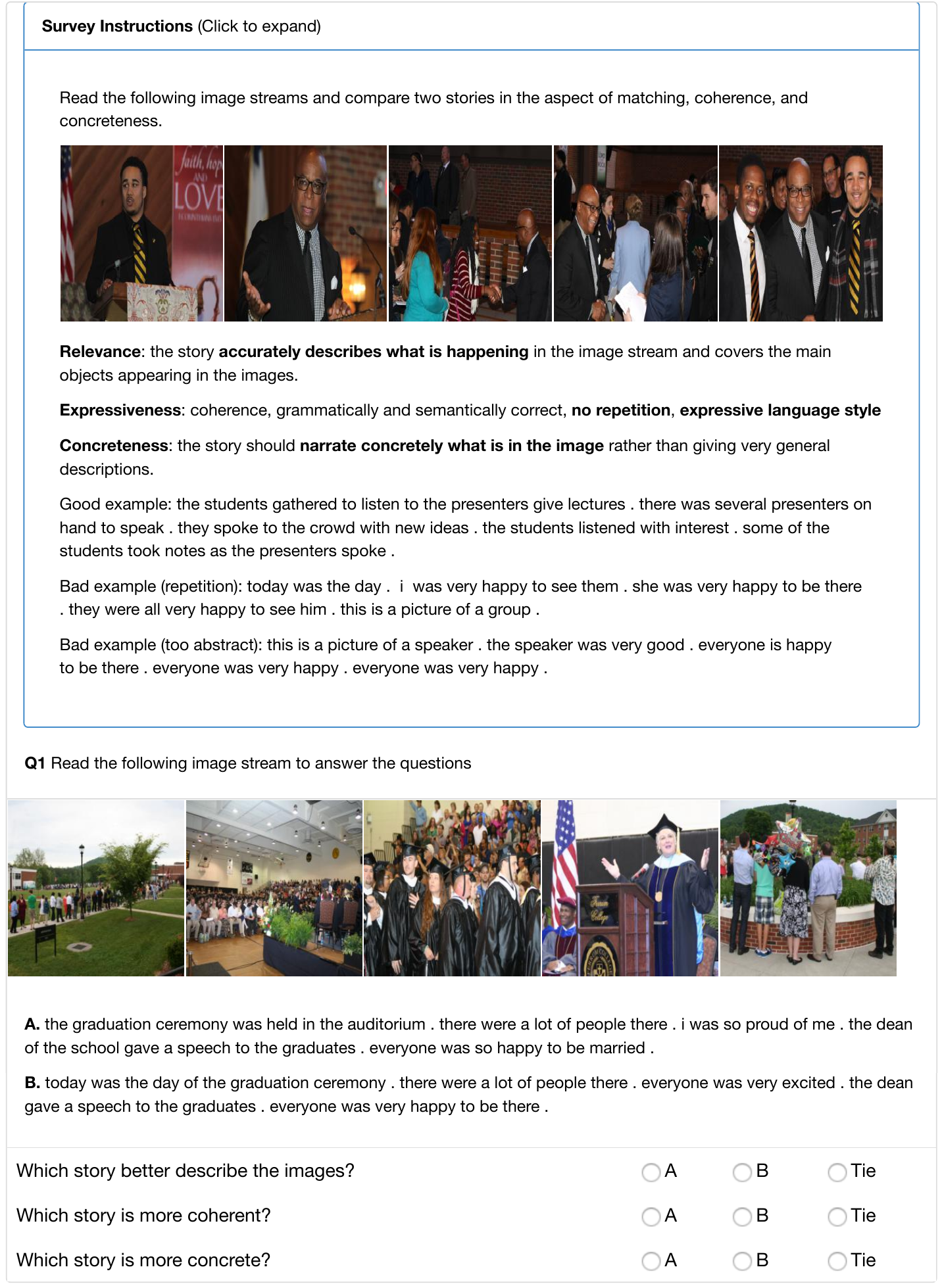}      
\caption{Pairwise Comparison Form}
\label{fig:compare}
\end{figure*}

We used AMT to perform two surveys, one picks a more human-like story. We asked the worker to answers 8 questions within 30 minutes, and we pay 5 workers to work on the same sheet to eliminate human-to-human bias. Here we demonstrate the Turing survey form in~\autoref{fig:turing}. Besides, we also perform a head-to-head comparison with other algorithms, we demonstrate the survey form in~\autoref{fig:compare}.

\end{document}